\documentclass[nohyperref]{article}

\usepackage{microtype}
\usepackage{graphicx}
\usepackage{booktabs} 
\usepackage{enumitem}
\usepackage{subcaption}

\usepackage{hyperref}

\usepackage[accepted]{icml2022}

\usepackage{amsmath}
\usepackage{amssymb}
\usepackage{mathtools}
\usepackage{amsthm}

\usepackage[capitalize,noabbrev]{cleveref}

\theoremstyle{plain}

\theoremstyle{definition}

\theoremstyle{remark}

\newcommand{\widen}{\triangledown}

\usepackage[textsize=tiny]{todonotes}

\icmltitlerunning{Abstract Interpretation for Generalized Heuristic Search in Model-Based Planning}

\begin{document}

\twocolumn[
\icmltitle{Abstract Interpretation for \\ Generalized Heuristic Search in Model-Based Planning}



\icmlsetsymbol{equal}{*}

\begin{icmlauthorlist}
\icmlauthor{Tan Zhi-Xuan}{mit}
\icmlauthor{Joshua B. Tenenbaum}{mit}
\icmlauthor{Vikash K. Mansinghka}{mit}
\end{icmlauthorlist}

\icmlaffiliation{mit}{Massachussetts Institute of Technology}

\icmlcorrespondingauthor{Tan Zhi-Xuan}{xuan@mit.edu}

\icmlkeywords{Abstract Interpretation, Model-Based Planning, Heuristic Search, Programming Languages, Reasoning}

\vskip 0.3in
]



\printAffiliationsAndNotice{}  

\begin{abstract}
Domain-general model-based planners often derive their generality by constructing search heuristics through the relaxation or abstraction of symbolic world models. We illustrate how abstract interpretation can serve as a unifying framework for these abstraction-based heuristics, extending the reach of heuristic search to richer world models that make use of more complex datatypes and functions (e.g. sets, geometry), and even models with uncertainty and probabilistic effects.
\end{abstract}

\section{Introduction}

Since the advent of the Stanford Research Institute Problem Solver (STRIPS) \cite{fikes1971strips}, it has been understood that planning and sequential decision making can be viewed as a form of theorem proving, where sequences of actions are derived via heuristic search given a symbolic world model and goal. While similar formal approaches have subsequently been highly successful in model checking \cite{clarke1997model}, program analysis \cite{nielson2004principles}, and constraint satisfaction \cite{barrett2018satisfiability}, they have come to play less of a role in automated planning and decision making, in light of successful learning-based approaches. \cite{mnih2015human,silver2017mastering}. This is in part due to the difficulty of specifying accurate symbolic models of the world, given the presence of uncertainty and limited expressiveness, e.g., to models with only propositional variables \cite{jimenez2012review}. Nonetheless, symbolic methods remain widely used in planning and robotics, often in conjunction with machine learning or motion planning \cite{leonetti2016synthesis,garrett2021integrated}.

Here, we show one powerful way of making these symbolic planning methods more general: Using abstract interpretation \cite{cousot1977abstract} to construct generalized heuristics for search. Domain-general planners often construct search heuristics by planning in a \emph{relaxed} or \emph{abstracted} model: the cost of a solution in a relaxed model can be used as an (optimistic) estimate of the true cost, providing heuristic guidance in search algorithms. Some of the abstractions used by these heuristics are also used in model checking \cite{seipp2018counterexample} or numeric invariant analysis \cite{scala2016interval}. However, they have typically been limited to propositional variables, with a few numeric extensions. Inspired by work on semantic modularity for symbolic planners \cite{gregory2012planning}, we illustrate how abstract interpretation can serve as a unifying framework for these abstraction-based heuristics. This leads to natural extensions of heuristic search to richer world models defined using more complex datatypes (e.g. sets) and functions (e.g. geometric operations), or even models with uncertainty and probabilistic effects. These heuristics can also be integrated with learning, allowing agents to jumpstart planning in novel world models via abstraction-derived information that is later refined by experience. This suggests that abstract interpretation can play a key role in building universal reasoning systems.

\section{Planning in Symbolic World Models}

Symbolic world models, also called \emph{planning domains}, describe the transition dynamics of an environment in symbolic terms. A planning domain constitutes part of a \emph{planning problem}, which also specifies the initial state and goal. While domains and problems are usually specified in a first-order language such as PDDL \cite{mcdermott1998pddl}, we adopt a simplified formalism for ease of exposition. Here, a planning problem is a tuple $(F,V,A,I,G)$, where:
\begin{itemize}[itemsep=1pt,parsep=0pt,topsep=0pt]
    \item $F$ is a set of functions;
    \item $V$ is a set of (typed) variables which may have Boolean, numeric, or other datatypes;
    \item $A$ is a set of actions $a \in A$, where:
    \begin{itemize}
        \item $\texttt{pre}(a)$ is a precondition for $a$, comprising Boolean functions from $F$ over variables in $V$;
        \item $\texttt{eff}(a)$ is the effect of $a$, a set of non-conflicting assignments to variables in $v \in V$
    \end{itemize}
    \item $I$ is the initial state, a set of assignments to each $v \in V$;
    \item $G$ is the goal, a logical formula of Boolean functions from $F$ over variables in $V$
\end{itemize}

\subsection{Action Semantics}

\begin{figure}[t]
    \scriptsize
    \begin{align*}
     &\textbf{\texttt{board\_person1\_plane1\_city1}} \\
     \texttt{\textbf{pre:}} &\ \texttt{person1\_loc} = \texttt{city1\_loc} \land \texttt{plane1\_loc} = \texttt{city1\_loc} \\[-2pt]
     \texttt{\textbf{eff:}} &\ \texttt{person1\_loc} := \texttt{plane1\_id}; \\[-2pt]
                   &\ \texttt{onboard1} := \texttt{add\_elem(onboard1, person1\_id)} \\
     &\textbf{\texttt{fly\_city1\_city2}}\\
     \texttt{\textbf{pre:}} &\ \texttt{plane1\_loc} = \texttt{city1\_loc} \ \land \\[-2pt]
     &\ \texttt{fuel1} > \texttt{burn\_rate} * \texttt{cardinality(onboard1)} \\[-2pt]
     \texttt{\textbf{eff:}} &\ \texttt{plane1\_loc} := \texttt{city2\_loc}; \\[-2pt]
    &\ \texttt{fuel1} := \texttt{fuel1} - \texttt{burn\_rate} * \texttt{cardinality(onboard1)};
    \end{align*}
    \vspace{-25pt}
    \caption{Example actions in an air-travel problem, with finite-domain, numeric, and set-valued variables.}
    \label{fig:actions}
\end{figure}

Assignment can be assumed to follow the standard semantics for imperative languages. Predicates and functions used in planning domains typically include testing if a Boolean variable is false or true, along with arithmetic operations \cite{fox2003pddl2}, but can be defined for new datatypes as necessary \cite{gregory2012planning}. Since assignments in the effect $\texttt{eff}(a)$ of an action $a$ must be non-conflicting, $\texttt{eff}(a)$ corresponds to parallel execution of each assignment. We can hence specify a denotational semantics for $a$ by identifying it with a relation over states $(S, T) \in a$, where $\texttt{pre}(a) \models S$, and $T$ is the result of applying $\texttt{eff}(a)$ to $S$. This is equivalent to the semantics for atomic actions in concurrent programs by \citet{lamport1990win}. Alternatively, actions can be seen as (guarded) primitive commands in a domain specific imperative language, with semantics defined by condition-checking and assignment as sub-primitives. Two example actions are shown in Figure \ref{fig:actions}.

\subsection{Planning Algorithms}

Common approaches to solving symbolic planning problems include backward search from the goal $G$, or forward best-first search from the initial state $I$ \cite{bonet2001planning}, with the fastest planners typically relying on variants of forward A* search guided by highly informative heuristics \cite{helmert2006fast}, along with compilation techniques for speed and memory efficiency \cite{helmert2009concise,zhixuan2022pddljl}. Heuristic-guided search can also be extended to probabilistic and partially observable domains, using algorithms such as Real Time Dynamic Programming (RTDP) \cite{bonet2003labeled} and Trial-based Heuristic Tree Search (THTS) \cite{keller2013trial}.

\section{Abstract Semantics for Symbolic Planning}

In order to perform abstract interpretation over symbolic world models, we need an abstraction function $\alpha(\cdot)$ that maps a concrete set of states $\mathcal{S}$ to an abstract state $\mathcal{S}^\sharp$, along with a concretization function $\gamma(\cdot)$ s.t. $\mathcal{S} \subseteq \gamma(\alpha(\mathcal{S}))$, where each concrete state $S \in \mathcal{S}$ is a complete assignment of values to the variables $V$ of the planning problem. In addition, we have to provide abstract semantics for each action, effectively abstract actions $a^\sharp$ that define a relation over abstract states $(\mathcal{S}^\sharp, \mathcal{T}^\sharp) \in a^\sharp$. We now describe several abstractions that are useful for deriving heuristics .

\subsection{State Abstractions}

\textbf{\emph{Cartesian Abstraction.}} One approach to abstracting states is to assign an abstract value to each variable $v \in V$, representing a set of concrete values $v$ could take. For example, the numeric variable \texttt{fuel1} in Figure \ref{fig:actions} could be assigned the interval value $[0, 1000)$. Doing this for each variable produces a \emph{Cartesian abstraction}: Given the values in an abstract state $\mathcal{S}^\sharp$, the corresponding set of concrete states $\gamma(\mathcal{S}^\sharp)$ is just the Cartesian product of sets associated with each abstract value \cite{ball2001boolean}.

\textbf{\emph{Predicate Abstraction.}} Another approach is to consider the set of all predicates $\mathcal{P}$ that feature in action preconditions $\texttt{pre}(a)$ for all $a \in A$, or the goal specification $G$. We define abstract states to be sets of these predicates $\mathcal{S}^\sharp \subseteq \mathcal{P}$, corresponding to the set of concrete states where every predicate in the abstract state holds true. The advantage of this abstraction is that it can be easily used to determine if an action is executable, of if the goal is achieved.

\subsection{Action Abstractions}

\begin{figure*}[t]
    \captionsetup[subfigure]{justification=centering}    \centering
    \begin{subfigure}[b]{\textwidth}
    \centering
    \includegraphics[width=0.75\textwidth]{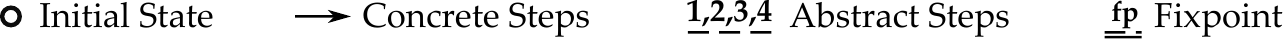}
    \vspace{6pt}  
    \end{subfigure}
    \begin{subfigure}[b]{0.45\textwidth}
    \centering
    \includegraphics[width=0.85\textwidth]{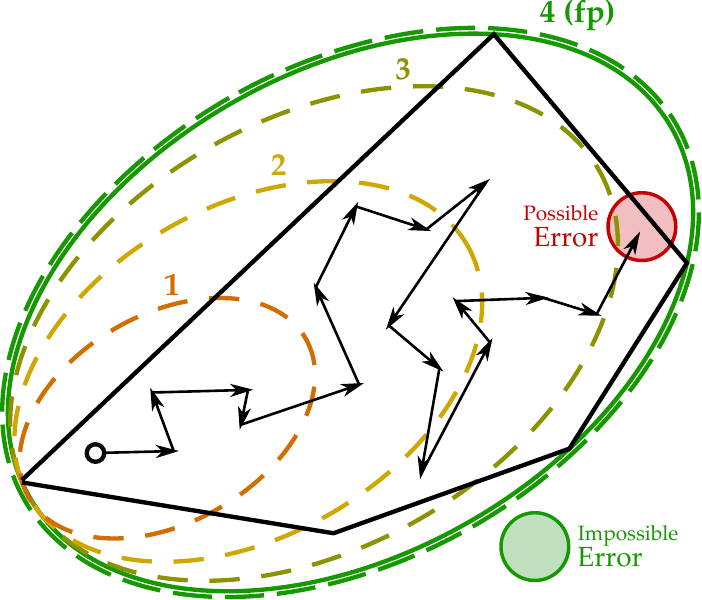}
    \caption{Abstract Interpretation for \\ Error Detection and Formal Verification}
    \end{subfigure}
    \begin{subfigure}[b]{0.45\textwidth}
    \centering
    \includegraphics[width=0.85\textwidth]{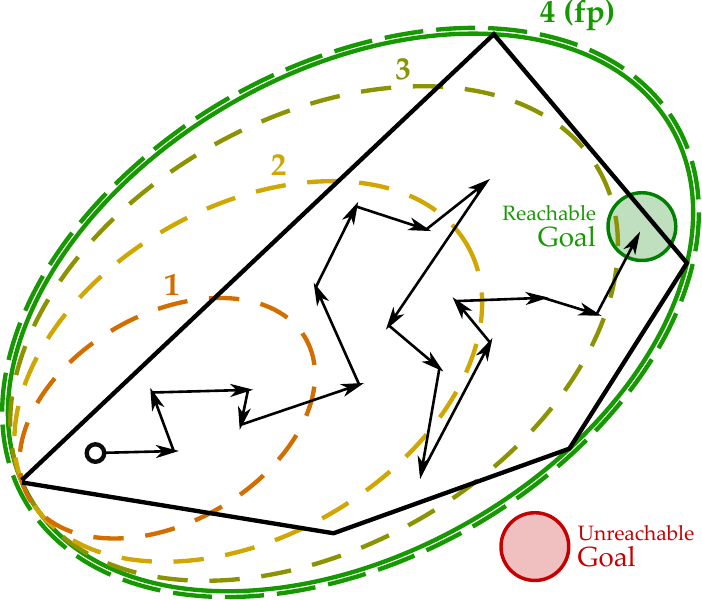}
    \caption{Abstract Planning for \\ Goal Reachability and Heuristic Estimation}
    \end{subfigure}
    \caption[Graphical analogy between abstract interpretation and abstract planning.]{A graphical analogy between (a) abstract interpretation for error detection and (b) abstract planning for goal reachability. In abstract interpretation, over-approximating the set of reachable states via abstract steps allows us to rule out impossible errors. In abstract planning, we can similarly rule out some goals as unreachable, or otherwise estimate how many steps it takes to reach them.}
    \label{fig:absint-analogy}
\end{figure*}

\textbf{\emph{State-Induced.}} Given abstraction and concretization functions $\alpha$ and $\gamma$, this induces a corresponding abstract action $a^\sharp$ for each action $a$: Given an abstract state $\mathcal{S}^\sharp$, we have $(\mathcal{S}^\sharp, \mathcal{T}^\sharp) \in a^\sharp$ where $\mathcal{T}^\sharp := \alpha(\{T\ |\ S \in \gamma(\mathcal{S}^\sharp), (S, T) \in a\})$. This can be efficiently implemented for Cartesian abstractions when the right-hand-side of all assignments in an action is constant, since the result of applying such an action is a specific concrete state which can then be abstracted.

\textbf{\emph{Widening-Based.}} Given a Cartesian state abstraction, a more relaxed abstraction is to replace each assignment $v := E$ in a concrete action $a$ with $v := v \widen E$, where $E$ is an expression and $\widen$ is a choice of widening operator associated with the value abstraction for variable $v$ \cite{gregory2012planning}. In the case where $v$ is Boolean or has a finite domain, a natural abstraction for $v$ is to replace the domain of $v$ with its powerset (e.g., allowing a Boolean $v$ to be true and false at the same time), using the join operator $\sqcup$ as a widening. Other choices are necessary when $v$ is numeric and unbounded, e.g. delayed widening \cite{mine2017tutorial}.

Widening-based abstract actions can be directly applied to Cartesian state abstractions, since they modify each state variable independently. They can also be extended to combined Cartesian and predicate abstractions, where each abstract state $\mathcal{S}^\sharp = (X^\sharp, P)$ comprises both abstract assignments $X^\sharp$ and predicates $P$. The successor of applying some action $a^\sharp$ is then $(\widen_a X^\sharp, Q)$, where $\widen_a X^\sharp$ is the widened set of assignments, and $Q$ is the largest set of predicates in $\mathcal{P}$ which hold true for at least some concrete $s \in \gamma(\widen_a X^\sharp)$. Note that $P \subseteq Q$ by construction (widening can only cause more predicates to be true), and $Q$ can be computed efficiently from $P$ and $\texttt{eff}(a^\sharp)$ by checking and adding only the predicates in $\mathcal{P}$ that are affected by $\texttt{eff}(a^\sharp)$.

\section{Deriving Heuristics via Abstraction}

Having described several abstractions for symbolic world models, we now illustrate how they can be used to derive multiple families of planning heuristics developed in the literature. We discuss novel extensions along the way.

\subsection{Relaxed Reachability via Widening}

As shown in Figure \ref{fig:absint-analogy}, determining the reachability of a goal condition in symbolic planning can be viewed as analogous to determining the reachability of an error in program verification. This analysis can be performed by imagining a non-deterministic search procedure that repeatedly applies all executable actions at once, then over-approximating that search process by merging the results of all actions. As noted by \citet{gregory2012planning}, this is equivalent to repeatedly applying widened versions of the actions until a goal condition or fixpoint is reached (though they do not describe it in terms of abstract interpretation). The number of iterations required serves as a lower bound on the true number of steps to the goal, and hence an optimistic heuristic estimate that can guide search. Indeed, when all variables are propositional, this is equivalent to the $h_\text{max}$ delete relaxation heuristic introduced by \citet{bonet2001planning}.

One issues with this analysis is that it may not terminate if variables have infinite domains and the widening operator $\widen$ is not appropriately chosen. To address this, \citet{gregory2012planning} employ an equivalent of delayed widening. However, many other widening strategies have been used in abstract interpretation, such as widening with thresholds \cite{mine2017tutorial}, that could lead to improved trade-offs between informativeness and convergence time when computing the heuristic.

\subsection{Relaxed Reachability over Predicate Sets}

An alternative formulation of the widening-based reachability analysis is to consider planning over \emph{predicate sets} using the predicate abstraction described earlier. More specifically, consider the (abstract) state space graph induced by the combined Cartesian and predicate abstraction, with states $\mathcal{S}^\sharp = (X^\sharp, P)$. Following \citet{haslum2000admissible}, finding the shortest path to a goal-satisfying state in the original problem is equivalent to finding the shortest path to an abstract state $(X^\sharp, P)$ where $P = G$ (we assume w.l.o.g. that the goal $G$ is a conjunction of predicates).

Now we perform several relaxations on this state space graph. First, we add edges corresponding to the widened actions described earlier, such that $(X^\sharp, P)$ connects to $(\widen_a X^\sharp, Q)$ for a widened action $a^\sharp$. Next, for every node $(X^\sharp, P)$ that has $(\widen_a X^\sharp, Q)$ as a successor, we add an edge from $(X^\sharp, P)$ to each node $(\cdot, R)$ where $Q \to R$. Finally, for each node $(X^\sharp, Q)$, we add a zero-cost edge \emph{from} every single-predicate node $(\cdot, R)$ s.t. $R \subseteq Q$ and $|R| = 1$. These relaxations achieve the following facts:

\begin{itemize}
    \item The cost of the shortest path in the relaxed graph $C$ is now less than the cost of the shortest path in the original problem $C^*$.
    
    \item By the second relaxation, we can reach every single-predicate node $(\cdot, R)$, $|R| = 1$ with less than or equal cost to reaching a multi-predicate node.
    
    \item By the third relaxation, we can compute the cost of reaching any multi-predicate node $C(\cdot, Q)$ as the max over costs of achieving each predicate in that node:
    $$C(\cdot, Q) = \max_{R \subseteq Q, |R| = 1} C(\cdot, R)$$
\end{itemize}

When all variables $V$ are propositional, and the right hand sides of all assignments are constant, this again recovers the $h_\text{max}$ heuristic, reformulated as the the $h^1$ heuristic in \citet{haslum2000admissible}. However, by incorporating the widening-based abstraction from earlier, this generalizes the heuristic to numeric variables, or any other datatype with an effective abstraction (e.g. approximating geometric sets with zonotopes \cite{bogomolov2019juliareach}). Furthermore, this derivation allows for a number of simple modifications, such as computing the cost of achieving a multi-predicate node as the \emph{sum} instead of the maximum over the costs of each individual predicate. This gives a generalized version of the widely used additive heuristic $h_\text{add}$, subsuming existing work on numeric subgoaling heuristics \cite{scala2020subgoaling}. While the result heuristic is non-admissible, it is generally much more informative and useful for search. 

\subsection{Counterexample Guided Abstraction Refinement}

A final class of abstraction heuristics enabled by abstract interpretation is counter-example guided abstraction refinement (CEGAR) \cite{seipp2018counterexample}. Starting with a coarse-grained Cartesian abstraction, one can solve for abstract plans to estimate the distance to the goal, and then iteratively refine the abstraction when counterexamples to the correctness of the abstraction are found. Each time after refinement, heuristic values obtained from the previous iteration can be used to guide search in the current iteration, leading to a rapid iterative approach that automatically constructs useful abstractions for planning.

However, CEGAR heuristics have so far been limited to problems with propositional variables, and CEGAR-like algorithms have only recently been explored in symbolic-geometric contexts for task and motion planning \cite{thomason2021counterexample}. By leveraging abstract semantics introduced for other datatypes in program analysis, it may be possible to extend the reach of such heuristics to a broader array of problems and domains.

\section{Other Extensions and Applications}

Thus far, we have focused on how abstract interpretation can be used to derive more general heuristics for forward search. However, there are many other possible applications of abstract interpretation, which we briefly reflect upon here.

\subsection{Uncertainty and Learning}

As noted earlier, heuristics can be used within algorithms such as Real Time Dynamic Programming (RTDP) \cite{bonet2003labeled} and Trial-based Heuristic Tree Search (THTS) \cite{keller2013trial} that operate over stochastic domains, using them to initialize an estimate of the value function for the underlying Markov Decision Process. To derive these heuristics from formal analysis of domains with probabilistic effects, the simplest abstraction would be to assume that all branches of a (discrete) probabilistic choice occur at once. This would effectively relax the problem, enabling the computation of admissible heuristics that guarantee eventual convergence to the true value function \cite{barto1995learning}. For continuous probability distributions with bounded support, it would also be possible to abstract distributions with their upper and lower bounds.

Using abstraction heuristics to initialize value functions is closely related to the idea of using heuristics as a learning target for neural network estimators of expected value \cite{shen2020learning,gehring2021reinforcement}. This would allow agents to jumpstart planning in novel world models via abstraction-derived information that is later refined by experience.

\subsection{Reverse Interpretation for Bidirectional Search}

Abstract interpretation can be executed in reverse \cite{hughes1987backwards,monniaux2001backwards}, allowing for generalizations of backward search and heuristic construction to domains with non-propositional variables. This might allow for novel combinations of (abstract) backwards search with (concrete) forward search, as has been explored in robotics \cite{garrett2018ffrob} and program analysis \cite{dinges2014targeted}.

\subsection{Abstract Interpretation for Generalized Planning}

Generalized plans, or policies, often take the form of imperative programs with control flow \cite{andre2002state,jimenez2019review,segovia2021generalized}. This suggests that abstraction-guided program synthesis methods \cite{solar2008program,srivastava2010program,wang2017program} can also be used for generalized planning.

This extended abstract presents only an initial foray into the manifold connections between abstract interpretation (AI) and artificially intelligent (AI) planning. We hope that illustrating some of these connections lays the ground for future interdisciplinary work.

\bibliography{paper}

\begin{thebibliography}{41}
\providecommand{\natexlab}[1]{#1}
\providecommand{\url}[1]{\texttt{#1}}
\expandafter\ifx\csname urlstyle\endcsname\relax
  \providecommand{\doi}[1]{doi: #1}\else
  \providecommand{\doi}{doi: \begingroup \urlstyle{rm}\Url}\fi

\bibitem[Andre \& Russell(2002)Andre and Russell]{andre2002state}
Andre, D. and Russell, S.~J.
\newblock State abstraction for programmable reinforcement learning agents.
\newblock In \emph{Aaai/iaai}, pp.\  119--125, 2002.

\bibitem[Ball et~al.(2001)Ball, Podelski, and Rajamani]{ball2001boolean}
Ball, T., Podelski, A., and Rajamani, S.~K.
\newblock Boolean and cartesian abstraction for model checking c programs.
\newblock In \emph{International Conference on Tools and Algorithms for the
  Construction and Analysis of Systems}, pp.\  268--283. Springer, 2001.

\bibitem[Barrett \& Tinelli(2018)Barrett and
  Tinelli]{barrett2018satisfiability}
Barrett, C. and Tinelli, C.
\newblock Satisfiability modulo theories.
\newblock In \emph{Handbook of model checking}, pp.\  305--343. Springer, 2018.

\bibitem[Barto et~al.(1995)Barto, Bradtke, and Singh]{barto1995learning}
Barto, A.~G., Bradtke, S.~J., and Singh, S.~P.
\newblock Learning to act using real-time dynamic programming.
\newblock \emph{Artificial intelligence}, 72\penalty0 (1-2):\penalty0 81--138,
  1995.

\bibitem[Bogomolov et~al.(2019)Bogomolov, Forets, Frehse, Potomkin, and
  Schilling]{bogomolov2019juliareach}
Bogomolov, S., Forets, M., Frehse, G., Potomkin, K., and Schilling, C.
\newblock Juliareach: a toolbox for set-based reachability.
\newblock In \emph{Proceedings of the 22nd ACM International Conference on
  Hybrid Systems: Computation and Control}, pp.\  39--44, 2019.

\bibitem[Bonet \& Geffner(2001)Bonet and Geffner]{bonet2001planning}
Bonet, B. and Geffner, H.
\newblock Planning as heuristic search.
\newblock \emph{Artificial Intelligence}, 129\penalty0 (1-2):\penalty0 5--33,
  2001.

\bibitem[Bonet \& Geffner(2003)Bonet and Geffner]{bonet2003labeled}
Bonet, B. and Geffner, H.
\newblock Labeled {RTDP}: Improving the convergence of real-time dynamic
  programming.
\newblock In \emph{ICAPS}, volume~3, pp.\  12--21, 2003.

\bibitem[Clarke(1997)]{clarke1997model}
Clarke, E.~M.
\newblock Model checking.
\newblock In \emph{International Conference on Foundations of Software
  Technology and Theoretical Computer Science}, pp.\  54--56. Springer, 1997.

\bibitem[Cousot \& Cousot(1977)Cousot and Cousot]{cousot1977abstract}
Cousot, P. and Cousot, R.
\newblock Abstract interpretation: a unified lattice model for static analysis
  of programs by construction or approximation of fixpoints.
\newblock In \emph{Proceedings of the 4th {ACM} {SIGACT-SIGPLAN} {S}ymposium on
  {P}rinciples of {P}rogramming {L}anguages}, pp.\  238--252, 1977.

\bibitem[Dinges \& Agha(2014)Dinges and Agha]{dinges2014targeted}
Dinges, P. and Agha, G.
\newblock Targeted test input generation using symbolic-concrete backward
  execution.
\newblock In \emph{Proceedings of the 29th ACM/IEEE international conference on
  Automated software engineering}, pp.\  31--36, 2014.

\bibitem[Fikes \& Nilsson(1971)Fikes and Nilsson]{fikes1971strips}
Fikes, R.~E. and Nilsson, N.~J.
\newblock {STRIPS}: A new approach to the application of theorem proving to
  problem solving.
\newblock \emph{Artificial intelligence}, 2\penalty0 (3-4):\penalty0 189--208,
  1971.

\bibitem[Fox \& Long(2003)Fox and Long]{fox2003pddl2}
Fox, M. and Long, D.
\newblock {PDDL}2. 1: An extension to {PDDL} for expressing temporal planning
  domains.
\newblock \emph{Journal of artificial intelligence research}, 20:\penalty0
  61--124, 2003.

\bibitem[Garrett et~al.(2018)Garrett, Lozano-Perez, and
  Kaelbling]{garrett2018ffrob}
Garrett, C.~R., Lozano-Perez, T., and Kaelbling, L.~P.
\newblock Ffrob: Leveraging symbolic planning for efficient task and motion
  planning.
\newblock \emph{The International Journal of Robotics Research}, 37\penalty0
  (1):\penalty0 104--136, 2018.

\bibitem[Garrett et~al.(2021)Garrett, Chitnis, Holladay, Kim, Silver,
  Kaelbling, and Lozano-P{\'e}rez]{garrett2021integrated}
Garrett, C.~R., Chitnis, R., Holladay, R., Kim, B., Silver, T., Kaelbling,
  L.~P., and Lozano-P{\'e}rez, T.
\newblock Integrated task and motion planning.
\newblock \emph{Annual review of control, robotics, and autonomous systems},
  4:\penalty0 265--293, 2021.

\bibitem[Gehring et~al.(2021)Gehring, Asai, Chitnis, Silver, Kaelbling,
  Sohrabi, and Katz]{gehring2021reinforcement}
Gehring, C., Asai, M., Chitnis, R., Silver, T., Kaelbling, L.~P., Sohrabi, S.,
  and Katz, M.
\newblock Reinforcement learning for classical planning: Viewing heuristics as
  dense reward generators.
\newblock \emph{arXiv preprint arXiv:2109.14830}, 2021.

\bibitem[Gregory et~al.(2012)Gregory, Long, Fox, and Beck]{gregory2012planning}
Gregory, P., Long, D., Fox, M., and Beck, J.~C.
\newblock Planning modulo theories: Extending the planning paradigm.
\newblock In \emph{Twenty-Second International Conference on Automated Planning
  and Scheduling}, 2012.

\bibitem[Haslum \& Geffner(2000)Haslum and Geffner]{haslum2000admissible}
Haslum, P. and Geffner, H.
\newblock Admissible heuristics for optimal planning.
\newblock In \emph{The Fifth International Conference on Artificial
  Intelligence Planning and Scheduling (AIPS), Breckenridge, Colorado, 14-17
  April, 2000.}, pp.\  140--149. AAAI Press, 2000.

\bibitem[Helmert(2006)]{helmert2006fast}
Helmert, M.
\newblock The {F}ast {D}ownward planning system.
\newblock \emph{Journal of Artificial Intelligence Research}, 26:\penalty0
  191--246, 2006.

\bibitem[Helmert(2009)]{helmert2009concise}
Helmert, M.
\newblock Concise finite-domain representations for {PDDL} planning tasks.
\newblock \emph{Artificial Intelligence}, 173\penalty0 (5-6):\penalty0
  503--535, 2009.

\bibitem[Hughes(1987)]{hughes1987backwards}
Hughes, J.
\newblock \emph{Backwards analysis of functional programs}.
\newblock University of Glasgow. Department of Computing Science, 1987.

\bibitem[Jim{\'e}nez et~al.(2012)Jim{\'e}nez, De~La~Rosa, Fern{\'a}ndez,
  Fern{\'a}ndez, and Borrajo]{jimenez2012review}
Jim{\'e}nez, S., De~La~Rosa, T., Fern{\'a}ndez, S., Fern{\'a}ndez, F., and
  Borrajo, D.
\newblock A review of machine learning for automated planning.
\newblock \emph{The Knowledge Engineering Review}, 27\penalty0 (4):\penalty0
  433--467, 2012.

\bibitem[Jim{\'e}nez et~al.(2019)Jim{\'e}nez, Segovia-Aguas, and
  Jonsson]{jimenez2019review}
Jim{\'e}nez, S., Segovia-Aguas, J., and Jonsson, A.
\newblock A review of generalized planning.
\newblock \emph{The Knowledge Engineering Review}, 34, 2019.

\bibitem[Keller \& Helmert(2013)Keller and Helmert]{keller2013trial}
Keller, T. and Helmert, M.
\newblock Trial-based heuristic tree search for finite horizon {MDP}s.
\newblock In \emph{Twenty-Third International Conference on Automated Planning
  and Scheduling}, 2013.

\bibitem[Lamport(1990)]{lamport1990win}
Lamport, L.
\newblock win and sin: Predicate transformers for concurrency.
\newblock \emph{ACM Transactions on Programming Languages and Systems
  (TOPLAS)}, 12\penalty0 (3):\penalty0 396--428, 1990.

\bibitem[Leonetti et~al.(2016)Leonetti, Iocchi, and
  Stone]{leonetti2016synthesis}
Leonetti, M., Iocchi, L., and Stone, P.
\newblock A synthesis of automated planning and reinforcement learning for
  efficient, robust decision-making.
\newblock \emph{Artificial Intelligence}, 241:\penalty0 103--130, 2016.

\bibitem[McDermott et~al.(1998)McDermott, Ghallab, Howe, Knoblock, Ram, Veloso,
  Weld, and Wilkins]{mcdermott1998pddl}
McDermott, D., Ghallab, M., Howe, A., Knoblock, C., Ram, A., Veloso, M., Weld,
  D., and Wilkins, D.
\newblock {PDDL} - the {P}lanning {D}omain {D}efinition {L}anguage, 1998.

\bibitem[Min{\'e}(2017)]{mine2017tutorial}
Min{\'e}, A.
\newblock Tutorial on static inference of numeric invariants by abstract
  interpretation.
\newblock \emph{Foundations and Trends in Programming Languages}, 4\penalty0
  (3-4):\penalty0 120--372, 2017.

\bibitem[Mnih et~al.(2015)Mnih, Kavukcuoglu, Silver, Rusu, Veness, Bellemare,
  Graves, Riedmiller, Fidjeland, Ostrovski, et~al.]{mnih2015human}
Mnih, V., Kavukcuoglu, K., Silver, D., Rusu, A.~A., Veness, J., Bellemare,
  M.~G., Graves, A., Riedmiller, M., Fidjeland, A.~K., Ostrovski, G., et~al.
\newblock Human-level control through deep reinforcement learning.
\newblock \emph{nature}, 518\penalty0 (7540):\penalty0 529--533, 2015.

\bibitem[Monniaux(2001)]{monniaux2001backwards}
Monniaux, D.
\newblock Backwards abstract interpretation of probabilistic programs.
\newblock In \emph{European Symposium on Programming}, pp.\  367--382.
  Springer, 2001.

\bibitem[Nielson et~al.(2004)Nielson, Nielson, and
  Hankin]{nielson2004principles}
Nielson, F., Nielson, H.~R., and Hankin, C.
\newblock \emph{Principles of program analysis}.
\newblock Springer Science \& Business Media, 2004.

\bibitem[Scala et~al.(2016)Scala, Haslum, Thi{\'e}baux, and
  Ramirez]{scala2016interval}
Scala, E., Haslum, P., Thi{\'e}baux, S., and Ramirez, M.
\newblock Interval-based relaxation for general numeric planning.
\newblock In \emph{ECAI 2016}, pp.\  655--663. IOS Press, 2016.

\bibitem[Scala et~al.(2020)Scala, Haslum, Thi{\'e}baux, and
  Ramirez]{scala2020subgoaling}
Scala, E., Haslum, P., Thi{\'e}baux, S., and Ramirez, M.
\newblock Subgoaling techniques for satisficing and optimal numeric planning.
\newblock \emph{Journal of Artificial Intelligence Research}, 68:\penalty0
  691--752, 2020.

\bibitem[Segovia-Aguas et~al.(2021)Segovia-Aguas, Jim{\'e}nez, and
  Jonsson]{segovia2021generalized}
Segovia-Aguas, J., Jim{\'e}nez, S., and Jonsson, A.
\newblock Generalized planning as heuristic search.
\newblock In \emph{Proceedings of the International Conference on Automated
  Planning and Scheduling}, volume~31, pp.\  569--577, 2021.

\bibitem[Seipp \& Helmert(2018)Seipp and Helmert]{seipp2018counterexample}
Seipp, J. and Helmert, M.
\newblock Counterexample-guided cartesian abstraction refinement for classical
  planning.
\newblock \emph{Journal of Artificial Intelligence Research}, 62:\penalty0
  535--577, 2018.

\bibitem[Shen et~al.(2020)Shen, Trevizan, and Thi{\'e}baux]{shen2020learning}
Shen, W., Trevizan, F., and Thi{\'e}baux, S.
\newblock Learning domain-independent planning heuristics with hypergraph
  networks.
\newblock In \emph{Proceedings of the International Conference on Automated
  Planning and Scheduling}, volume~30, pp.\  574--584, 2020.

\bibitem[Silver et~al.(2017)Silver, Schrittwieser, Simonyan, Antonoglou, Huang,
  Guez, Hubert, Baker, Lai, Bolton, et~al.]{silver2017mastering}
Silver, D., Schrittwieser, J., Simonyan, K., Antonoglou, I., Huang, A., Guez,
  A., Hubert, T., Baker, L., Lai, M., Bolton, A., et~al.
\newblock Mastering the game of go without human knowledge.
\newblock \emph{nature}, 550\penalty0 (7676):\penalty0 354--359, 2017.

\bibitem[Solar-Lezama(2008)]{solar2008program}
Solar-Lezama, A.
\newblock \emph{Program synthesis by sketching}.
\newblock University of California, Berkeley, 2008.

\bibitem[Srivastava et~al.(2010)Srivastava, Gulwani, and
  Foster]{srivastava2010program}
Srivastava, S., Gulwani, S., and Foster, J.~S.
\newblock From program verification to program synthesis.
\newblock In \emph{Proceedings of the 37th annual ACM SIGPLAN-SIGACT symposium
  on Principles of programming languages}, pp.\  313--326, 2010.

\bibitem[Thomason \& Kress-Gazit(2021)Thomason and
  Kress-Gazit]{thomason2021counterexample}
Thomason, W. and Kress-Gazit, H.
\newblock Counterexample-guided repair for symbolic-geometric action
  abstractions.
\newblock \emph{arXiv preprint arXiv:2105.06537}, 2021.

\bibitem[Wang et~al.(2017)Wang, Dillig, and Singh]{wang2017program}
Wang, X., Dillig, I., and Singh, R.
\newblock Program synthesis using abstraction refinement.
\newblock \emph{Proceedings of the ACM on Programming Languages}, 2\penalty0
  (POPL):\penalty0 1--30, 2017.

\bibitem[Zhi-Xuan(2022)]{zhixuan2022pddljl}
Zhi-Xuan, T.
\newblock {PDDL}.jl: An extensible interpreter and compiler interface for fast
  and flexible {AI} planning.
\newblock Master's thesis, MIT, 2022.

\end{thebibliography}
\bibliographystyle{icml2022}

\end{document}